\pdfoutput=1
\PassOptionsToPackage{dvipsnames}{xcolor}
\PassOptionsToPackage{numbers}{natbib}
\documentclass[11pt]{article}

\usepackage[preprint]{acl}

\usepackage{times}
\usepackage{latexsym}
\usepackage[T1]{fontenc}
\usepackage[utf8]{inputenc}
\usepackage{microtype}
\usepackage{inconsolata}

\usepackage[dvipsnames]{xcolor}

\usepackage{graphicx}
\usepackage{booktabs} 
\usepackage{algorithm}
\usepackage{algorithmic}
\usepackage{tcolorbox}
\tcbuselibrary{skins}

\usepackage{mdframed}
\usepackage{enumitem}
\usepackage{multicol}
\usepackage{subfig}  
\usepackage{hyperref}
\usepackage{authblk}
\usepackage{geometry}
\usepackage[symbol]{footmisc}
\usepackage{caption}
\definecolor{Pink}{RGB}{245,215,205}
\definecolor{Thistle}{RGB}{216,191,216}

\title{
What's In Your Field? \\
Mapping Scientific Research with Knowledge Graphs and LLMs
}

\author[1,2]{\large Abhipsha Das}
\author[1,3]{\large Nicholas Lourie}
\author[1,3]{\large Siavash Golkar}
\author[1,4]{\large Mariel Pettee}

\affil[ ]{
\textsuperscript{1}Polymathic AI \hspace{0.3cm} 
\textsuperscript{2}Flatiron Institute \hspace{0.3cm} 
\textsuperscript{3}New York University \hspace{0.3cm}
\textsuperscript{4}Lawrence Berkeley National Laboratory
}

\affil[ ]{
\href{mailto:abhipsha.das@nyu.edu}{\texttt{abhipsha.das@nyu.edu}}
}

\begin{document}
\maketitle

\begin{abstract}
The scientific literature's exponential growth makes it increasingly challenging to navigate and synthesize knowledge across disciplines. Large language models (LLMs) are powerful tools for understanding scientific text, but they fail to capture detailed relationships across large bodies of work. Unstructured approaches, like retrieval augmented generation, can sift through such corpora to recall relevant facts; however, when millions of facts influence the answer, unstructured approaches become cost prohibitive. Structured representations offer a natural complement---enabling systematic analysis across the whole corpus. Recent work enhances LLMs with unstructured or semistructured representations of scientific concepts; to complement this, we try extracting structured representations using LLMs. By combining LLMs' semantic understanding with a schema of scientific concepts, we prototype a system that answers precise questions about the literature as a whole. Our schema applies across scientific fields and we extract concepts from it using only 20 manually annotated abstracts. To demonstrate the system, we extract concepts from 30,000 papers on arXiv spanning astrophysics, fluid dynamics, and evolutionary biology. The resulting database highlights emerging trends and, by visualizing the knowledge graph, offers new ways to explore the ever-growing landscape of scientific knowledge.\\
\textcolor{red}{Demo:} \href{https://huggingface.co/spaces/abby101/surveyor-0}{\emph{abby101/surveyor-0}} on HF Spaces. \textcolor{red}{Code:} \href{https://github.com/chiral-carbon/kg-for-science}{\emph{https://github.com/chiral-carbon/kg-for-science}}. 
\footnote[4]{\textcolor{red}{Video: \href{https://youtu.be/pNVULGb3gKo}{YouTube Link}}} \\
\end{abstract}

\section{Introduction}

Consider a researcher seeking to build a multimodal foundation model for astrophysics. They might begin by asking: What are the most important data modalities to support---the most common ones in the field? How would such a researcher go about answering these questions today?

One might hope that LLMs could easily answer such a question, but while they have created unprecedented opportunities for accelerating scientific discovery, they struggle to aggregate reliable statistics and generate systematic analyses across the breadth of scientific literature. Manually reviewing papers or consulting domain experts are not scalable approaches when there are thousands of papers to investigate.

Most current approaches rely on unstructured methods like retrieval augmented generation (RAG) \citep{lewis2021retrievalaugmentedgenerationknowledgeintensivenlp, lála2023paperqaretrievalaugmentedgenerativeagent, gao2024retrievalaugmentedgenerationlargelanguage}. While these methods excel at broad information retrieval and synthesis, they make it difficult to analyze specific patterns in research across large bodies of literature. The limitations become particularly evident when researchers try to understand how research in a field evolves. They need to track new instruments, identify research problems that require methodological innovation, and understand how theoretical models get validated across disciplines. Unstructured representations struggle to systematically capture these relationships.

Some efforts explore semistructured representations such as: keyphrase extractions based on statistical patterns \citep{gu2024interestingscientificideageneration} and LLM-based concept extraction combined with constructing vector-similarity-based knowledge graphs \citep{sun2024knowledgegraphastronomicalresearch}. While valuable, these methods typically treat all concepts uniformly without distinguishing between their functional roles in scientific work, or rely on semantic similarity of concepts using embedding models. This limits the utility of extracted knowledge when a researcher needs a quantitative analysis.

\begin{figure*}[ht]
\centering
\includegraphics[width=\textwidth, height=0.33\textwidth]{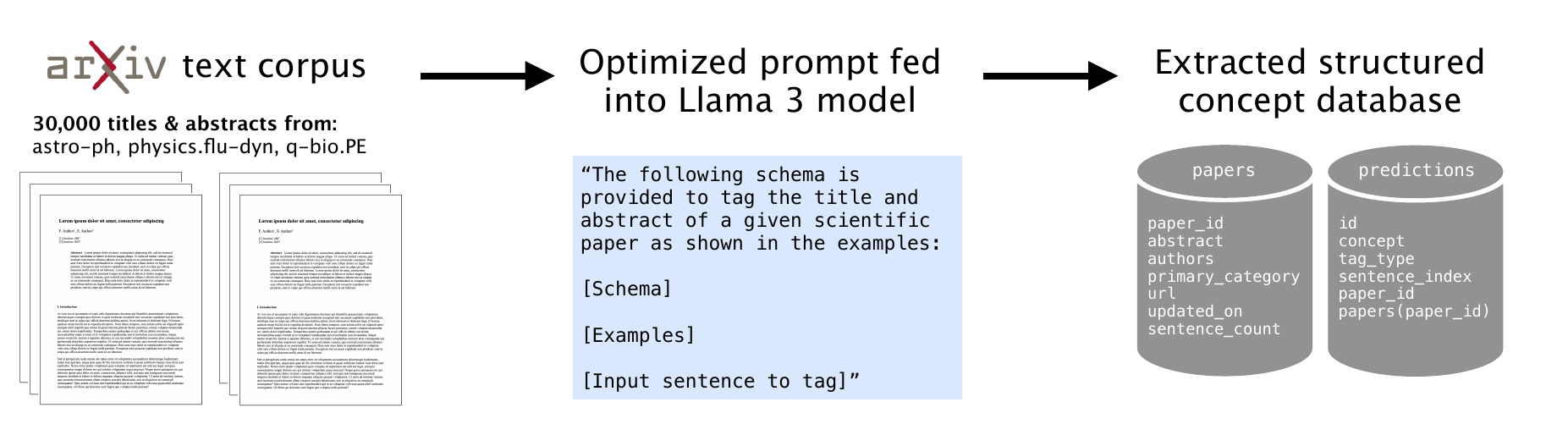}
\caption{Illustration of the structured concept extraction pipeline: i) the corpus used, ii) running optimized prompt on full corpus, iii) storing model’s outputs and corpus metadata in SQL database.}
\label{fig:pipeline}
\end{figure*}

To address this challenge, we introduce a novel approach using LLMs that extracts categorized concepts from scientific papers using a general schema covering key research entities like objects, datasets, methods, and modalities. Our system combines structured knowledge representation with an interactive query interface to enable researchers to analyze methodological patterns, track research evolution, and understand relationships between different aspects of scientific work at scale. We visualize the extracted structured information using knowledge graphs that provide key insights into concept co-occurrence in scientific research. The main contributions of this work are: 1) a generic schema for categorizing scientific concepts across different fields, 2) a scalable LLM-based extraction pipeline to mine concepts from papers, 3) an interactive system for querying scientific information, 4) informative knowledge graphs built from the extracted concepts to represent scientific fields.

\section{Method}
Our system consists of three key components: a \textit{schema} defining different kinds of scientific concepts, a \textit{pipeline} for prompting LLMs to extract these concepts, and a \textit{database} to store this structured knowledge for efficient queries and analysis.

\subsection{Schema} 
Through an iterative discussion process of example selection, comparing manual annotations between the authors, and examining scientific papers across different domains, we developed an annotation schema capturing nine fundamental categories of scientific concepts: \emph{models}, \emph{tasks}, \emph{datasets}, \emph{fields}, \emph{modalities}, \emph{methods}, \emph{objects}, \emph{properties}, and \emph{instruments}. In designing the schema, we aimed for categories that apply across scientific disciplines.

\begin{table}[ht]
\small
\begin{tabular}{@{}p{0.25\linewidth}@{}p{0.7\linewidth}@{}}
\texttt{model:} & representation of a (scientific) phenomenon using mathematical formalism and/or computational simulation \\
\texttt{task:} & specific problem, objective or goal to be accomplished \\
\texttt{dataset:} & collection of data, measurements or observations \\
\texttt{field:} & academic (sub)discipline \\
\texttt{modality:} & class or type of data/observations with similar or the same structure \\
\texttt{method:} & approach, technique or procedure to complete a task \\
\texttt{object:} & entity that can be studied \\
\texttt{property:} & quantitative or qualitative descriptor, or an inherent attribute of an entity, data, modality or method \\
\texttt{instrument:} & device or system used for making measurements \\
\end{tabular}
\caption{Definitions for a schema of scientific concepts.}
\label{schema_tab}
\end{table}

This schema intentionally uses broad category definitions that are immediately understandable to scientists without requiring study of specialized and intricate taxonomies. We opted for coarser categories to avoid the ambiguity and complexity that arises when making subtle distinctions between closely related concepts, even though this means some concepts could be described by multiple tags. Rather than implementing complex disambiguation tests, our simple tagging schema allows us to maintain scalability; however, there is always a trade-off between coverage and precision.

\begin{tcolorbox}[
    title=\textbf{Original Sentence},
    colback=Pink,
    colframe=Black!75,
    arc=3mm,
    boxrule=0.8pt,
    left=2mm,
    right=2mm,
    top=2mm,
    bottom=2mm,
    enhanced,
    drop shadow=Black
]
\noindent
We present an analysis of a new Australia Telescope Compact Array (ATCA) radio-continuum observation of supernova remnant (SNR) G1.9+0.3, which at an age of ~181±25 years is the youngest known in the Galaxy.
\end{tcolorbox}
\vspace{-6mm}
\begin{tcolorbox}[
    title=\textbf{Tagged Sentence},
    colback=Pink,
    colframe=Black!75,
    arc=3mm,
    boxrule=0.8pt,
    left=2mm,
    right=2mm,
    top=2mm,
    bottom=2mm,
    enhanced,
    drop shadow=Black
]
\noindent
\raggedright
We present an analysis of a new {\color{blue}\texttt{<dataset>}}
{\color{orange}\texttt{<instrument>}}Australia Telescope Compact 
Array (ATCA){\color{orange}\texttt{</instrument>}} {\color{green}\texttt{<modality>}}radio-
continuum{\color{green}\texttt{</modality>}} observation{\color{blue}\texttt{</dataset>}} 
of {\color{red}\texttt{<object>}}supernova remnant (SNR) G1.9+0.3
{\color{red}\texttt{</object>}}, which at an {\color{magenta}\texttt{<property>}}age{\color{magenta}\texttt{</property>}}
of ~181±25 years is the youngest known in 
the {\color{red}\texttt{<object>}}Galaxy{\color{red}\texttt{</object>}}.
\end{tcolorbox}

We implemented our extraction pipeline using the open-source Llama-3 70B Instruct model \citep{grattafiori2024llama3herdmodels}, employing few-shot learning to guide concept extraction. For our prompt optimization experiments, we manually annotated 20 papers, using 3 demonstration papers for few-shot examples and the remaining 17 as a development set to iteratively refine the prompts. Above is an example of the manual annotation process sentence-by-sentence.

The pipeline processes the language content sentence-by-sentence using manually annotated examples to demonstrate the target structure (see Fig.~\ref{fig:prompt_example}).

\begin{figure}[ht]
\centering
\begin{tcolorbox}[
    title=\textbf{Illustration: Prefix + Prompt},
    colback=white,
    colframe=blue!40,
    arc=0mm,
    boxrule=1pt,
    leftrule=3pt,
    left=3mm,
    right=3mm,
    top=3mm,
    bottom=3mm,
    enhanced,
    attach boxed title to top left={yshift=-2mm, xshift=3mm},
    boxed title style={
        colback=blue!40,
        colframe=blue!40,
        size=small,
        sharp corners
    }
]
The following schema is provided to tag the title and abstract of a given scientific paper as shown in the examples:

\vspace{1mm}
\$\hyperref[schema_tab]{SCHEMA}
\vspace{2mm}

\noindent
{\color{violet!80}\texttt{Sentence:}} This magnetic field strength implies a minimum total energy of the synchrotron radiation of E$_{\textrm{min}} \approx$ 1.8$\times$10$^{48}$ ergs.

{\color{violet!80}\texttt{Extractions:}}

\noindent
{\color{magenta!80}\texttt{property:}} magnetic field strength, energy\\
{\color{red!80}\texttt{object:}} synchrotron radiation

\vspace{2mm}
{\color{darkgray!80}\texttt{... (Total 9 few-shot examples) ...}}
\vspace{1mm}

\noindent
{\color{violet!80}\texttt{Sentence:}} We present HATNet observations of XO-5b, confirming its planetary nature based on evidence beyond that
described in the announcement of Burke et al
\end{tcolorbox}

\vspace{-6mm}

\begin{tcolorbox}[
    colback=white,
    colframe=blue!40,
    arc=0mm,
    boxrule=1pt,
    leftrule=3pt,
    left=3mm,
    right=3mm,
    top=3mm,
    bottom=3mm,
    enhanced,
    attach boxed title to top left={yshift=-2mm, xshift=3mm},
    boxed title style={
        colback=blue!40,
        colframe=blue!40,
        size=small,
        sharp corners
    }
]
\textbf{\underline{Ground Truth Tags:}}
\vspace{1mm}

\noindent
{\color{blue!80}\texttt{dataset:}} HATNet observations\\
{\color{orange!80}\texttt{instrument:}} HATNet\\
{\color{red!80}\texttt{object:}} XO-5b

\vspace{3mm}
\textbf{\underline{Predicted Tags:}}
\vspace{1mm}

\noindent
{\color{blue!80}\texttt{dataset:}} HATNet observations\\
{\color{red!80}\texttt{object:}} XO-5b, planetary nature
\end{tcolorbox}
\caption{Expanded prompt illustration with schema and few-shot examples, along with the sentence to predict tags for.}
\label{fig:prompt_example}
\end{figure}

\subsection{Pipeline}

To optimize extraction reliability, we conducted systematic prompt engineering experiments on the manually annotated set, varying: (i) number and selection of few-shot examples, (ii) structure and ordering of the prompt, (iii) granularity of input text (sentence vs. paragraph), (iv) format of extracted concepts (JSON vs. human-readable).

To guide this iteration, we used a comprehensive set of metrics calculated on our development set, including: precision, recall and F-1 scores, for exact matches. In addition, we also considered the processing time and efficiency of the different approaches. During the annotation process, we found that even simple and broad concepts, such as a modality, encounter ambiguities when applied across scientific fields. As a result, it is likely that no method can achieve complete agreement with the development set annotations. Rather than as an absolute benchmark, we used the development set as a directional signal---a way to see if a given change improved the extraction process. Ultimately, our optimized prompt configuration consisting of instruction, schema and 9 few-shot examples (3 sentences with annotated extractions from each of the 3 demonstration papers) shows promising results. The final results on our development set were: precision of 44\% $\pm$ 12\% and recall of 31\% $\pm$ 11\% in human-readable response format, with processing times averaging around 2.8 seconds per sentence. When using JSON output format, we observed similar precision levels with slightly higher recall rates of 40\% $\pm$ 12\% and processing times of around 4 seconds per sentence. While some noise persisted in the extracted data, the results were sufficient to explore using such structured knowledge from the scientific literature in order to discover relationships and systematically analyze complex statistical questions.

\subsection{Database}
The extracted concepts and their relationships are stored in a SQL database for scientific analysis queries which enables fast computation of aggregate statistics. The SQL database contains 2 tables: \texttt{papers} and \texttt{predictions} (see Fig.~\ref{fig:pipeline}). The \texttt{papers} table contains metadata, raw text, author and category information about the papers, and the \texttt{predictions} table contains the information about the tagged concepts, the tag type and the papers they come from. Storing both the extracted information and relevant metadata about where the extractions came from facilitates analyses such as the evolution of methodological approaches over time or the adoption patterns of new experimental techniques across fields.

\section{Demo}

\subsection{Visualization}
To explore the relationships between scientific concepts, we built a dynamic visualization using force-directed graph layouts. In this representation, nodes represent individual scientific concepts $V$ (e.g., specific methods, instruments, or objects of study), while edges represent co-occurrences within the same paper $E$ in a graph $G = (V, E)$. A physics-based spring layout algorithm determines the spatial arrangement, with frequently co-occurring concepts drawn closer together.

Our visualization system supports interactive exploration through: 1. Tag-type filtering to focus on specific concept categories (e.g., only methods or instruments), 2. Node highlighting to emphasize specific concepts and their immediate connections, 3. Depth-based exploration to reveal n-hop neighborhoods around concepts of interest, 4. Dynamic force-directed layout updates to reflect filtered subgraphs.

This graph-based approach enables both targeted investigation of specific concept relationships and broader analysis of methodological patterns across domains. For example, researchers can identify clusters of related experimental techniques, trace the adoption of methods across different subfields, or visualize isolated clusters of objects to understand how they are studied.

\subsection{Query Interface}
The query interface supports SQL queries for scientific concept exploration, with predefined queries demonstrating use cases like modality distribution analysis and temporal trends. The interface allows researchers to ask increasingly sophisticated questions by leveraging the structured database. For example, a researcher building an astrophysics foundation model could analyze most-used modalities, examine their current coverage, track usage trends over 5 years, and estimate coverage gains from adding new modalities. This approach helps scientists make data-driven research decisions that would be difficult to achieve through other means.

\section{Results}

\textbf{Dataset.} We collected the titles and abstracts from 30,000 articles from arXiv comprising 10,000 papers each from astrophysics, fluid dynamics and evolutionary biology, in order test across a breadth of scientific disciplines. Titles and abstracts (i.e. article metadata) were used instead of the full text in order to optimize processing efficiency while maintaining representativeness, since the titles and abstracts of papers are likely to be more information dense than the papers' bodies. After setting aside 20 astrophysics papers for prompt development, we used the optimized prompts refined through this development process to extract concepts from the remaining 9,980 astrophysics papers (from the original 10,000), as well as all 10,000 papers from each of the other two fields (fluid dynamics and evolutionary biology). Our final extractions and subsequent knowledge graph visualizations include results for these 29,980 papers.

\subsection{Graph-based Exploration}

\begin{figure*}[t]
    \centering
    \subfloat[Analysis of galaxy images: various galactic entities, measurement objects, clouds, instruments, and properties.]{\includegraphics[width=0.48\textwidth]{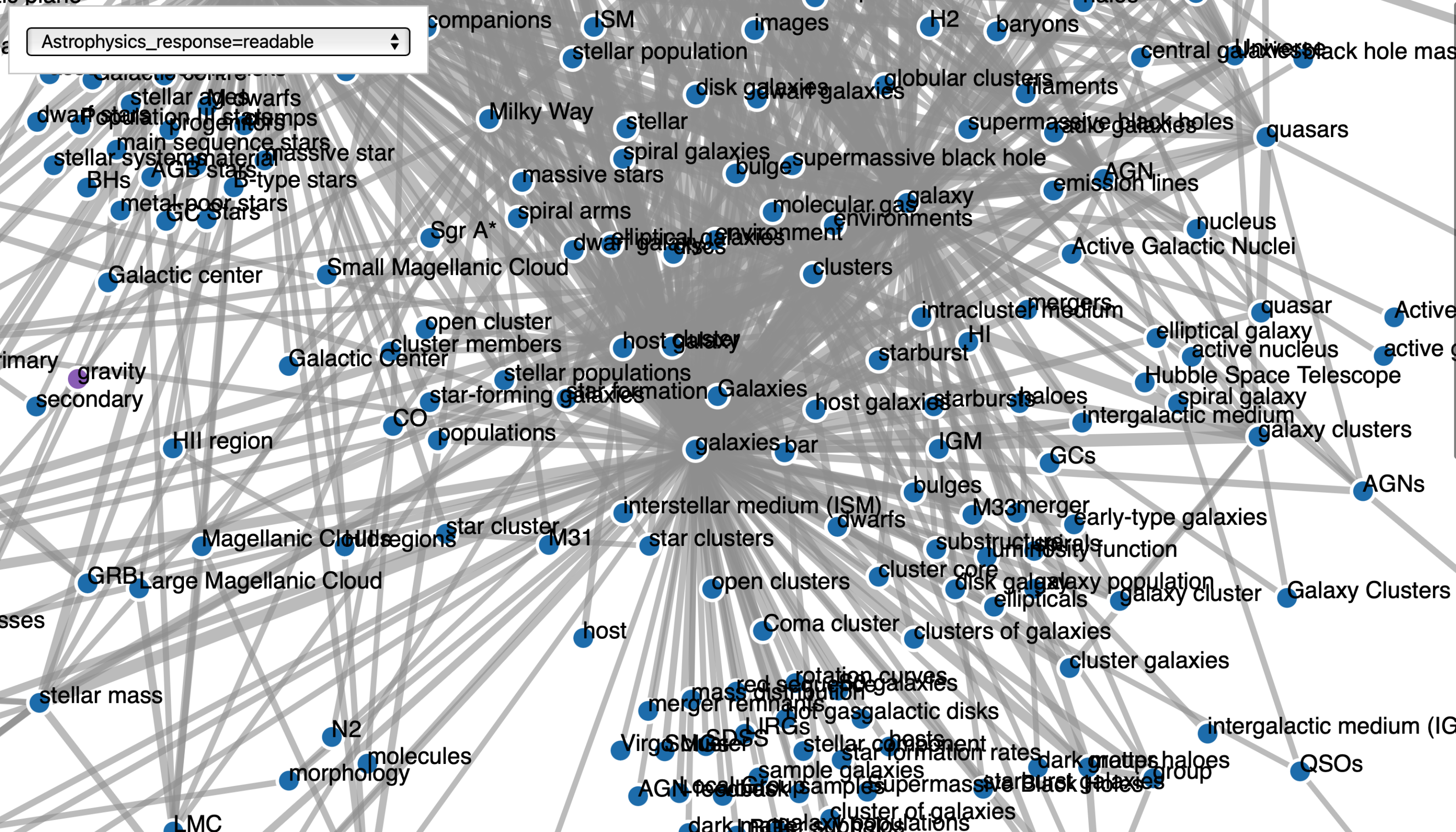}
    }
    \hfill
    \subfloat[COVID-19 cluster analysis: geographic distribution of pandemic research and immune health terms indicating research focus areas.]{
        \includegraphics[width=0.48\textwidth]{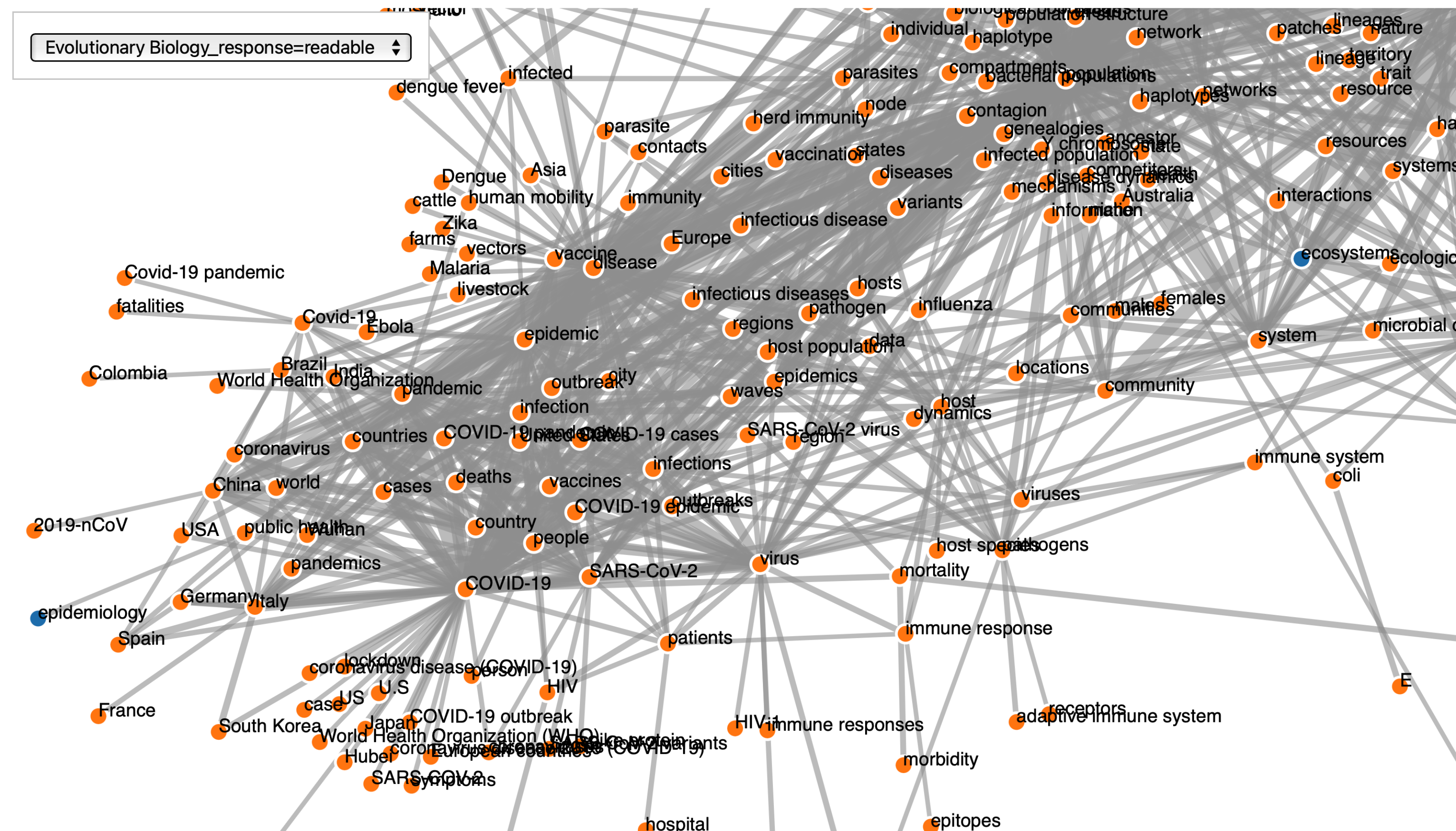}
    }

    \vspace{1mm}

    \subfloat[Fluid dynamics clusters centered around "flow," branching into related flow and turbulence-based physical phenomena.]{
        \includegraphics[width=0.48\textwidth]{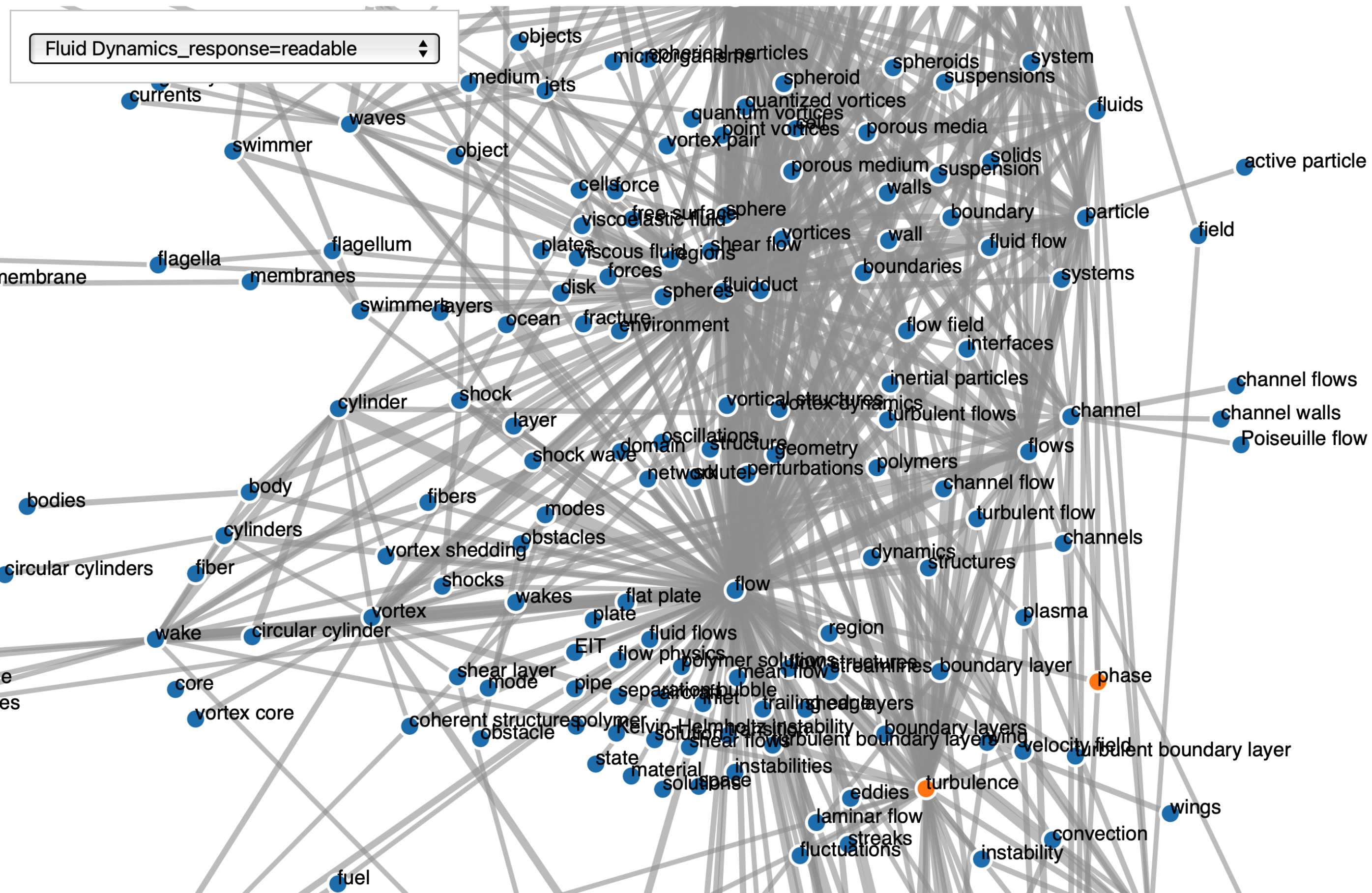}
    }
    \hfill
    \subfloat[Astrophysical objects and phenomena spanning multiple scales, from individual stars and binaries to galactic structures and clusters.]{
        \includegraphics[width=0.48\textwidth]{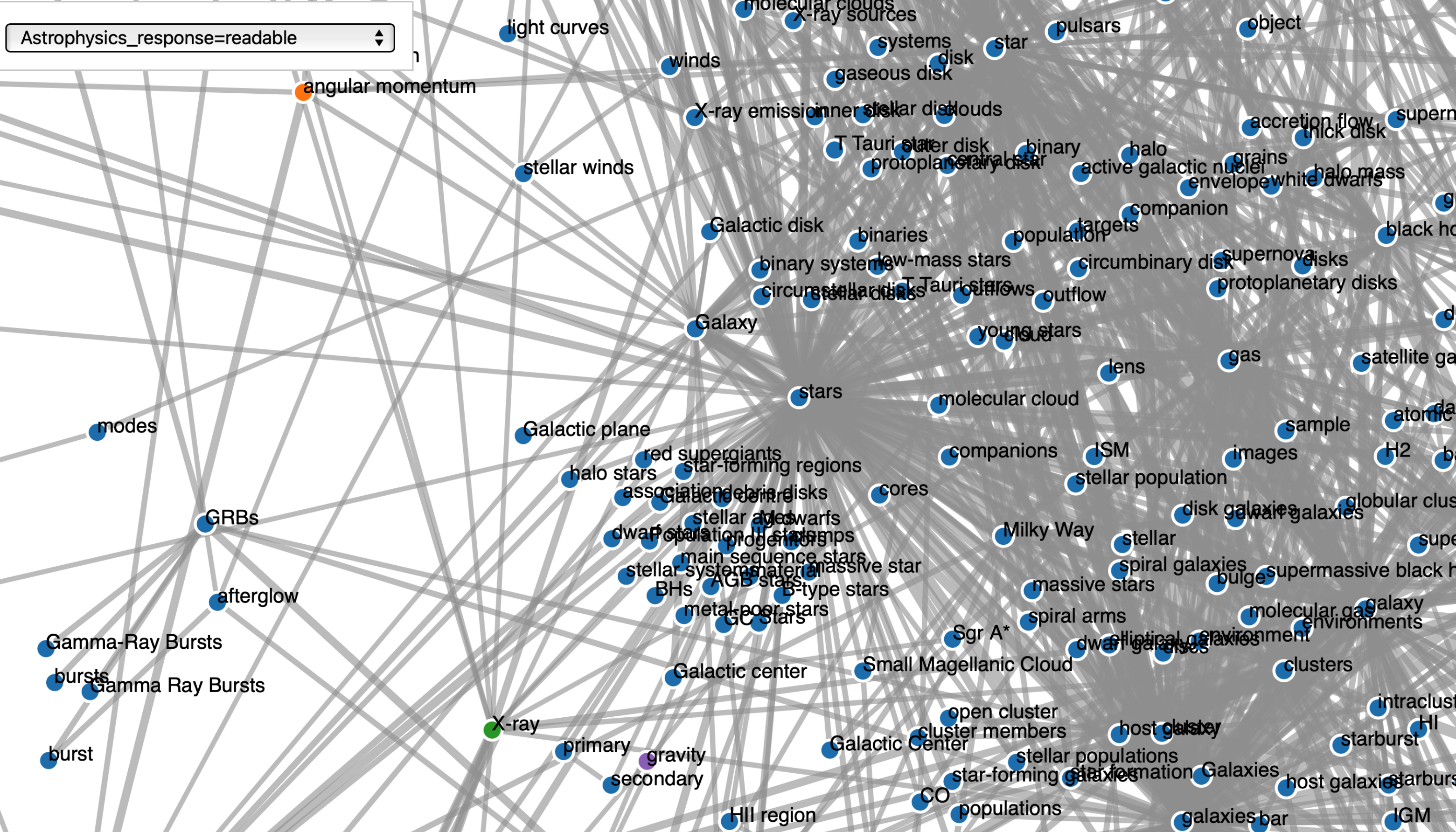}
    }

    \caption{Co-occurrence graphs: astrophysics (a, d), epidemiology (b), fluid dynamics (c).}
    \label{fig:fourgrid}
\end{figure*}

Fig. \ref{fig:fourgrid} demonstrates the interconnected nature of scientific concepts through co-occurrence knowledge graphs from our analyzed domains.\footnote[2]{We use the \href{https://d3js.org/d3-force}{d3-force} library; see documentation for more information on the spring-layout implementation.}

\subsection{Demonstrating Example Queries}
We examine several key questions to demonstrate the interface's capability for both exploratory research and targeted investigation in scientific discovery.

\vspace{3mm}
\pagebreak
\textbf{Temporal Evolution Analysis} 
\begin{mdframed}[
   linewidth=2pt,
   linecolor=gray!40,
   topline=false,
   rightline=false,
   bottomline=false,
   backgroundcolor=gray!5,
   innerleftmargin=10pt,
   innerrightmargin=10pt
]
   \textit{Example: What are the new datasets that came out in fluid dynamics since 2020?}

   \vspace{2mm}
   \begin{tabular}{p{0.3\linewidth}|p{0.3\linewidth}|p{0.2\linewidth}}
   \textbf{Paper} & \textbf{Dataset} & \textbf{Date} \\
   \hline
   Error propagation of direct pressure gradient inte... & Synthetic velocimetry data & Jul 22 2024 \\
   \hline
    Modelling turbulence in axisymmetric wakes: an app... & 3D laser Doppler anemometry data & Jul 12 2024 \\
   \end{tabular}
\end{mdframed}

Showing 2 of 524 results representing recent fluid dynamics datasets discussed in our corpus. Failure modes included retrievals like ``meteorological data'' and ``wind-tunnel data'', which are generic rather than specific datasets.

\vspace{10mm}

\textbf{Cross-disciplinary Method Application} 
 \vspace{2mm}
\begin{mdframed}[
    linewidth=2pt,
    linecolor=gray!40,
    topline=false,
    rightline=false,
    bottomline=false,
    backgroundcolor=gray!5,
    innerleftmargin=10pt,
    innerrightmargin=10pt
]
    \textit{Example: What are all the datasets in evolutionary biology that use PDEs?}

    \vspace{2mm}
    \begin{tabular}{p{0.3\linewidth}|p{0.25\linewidth}|p{0.2\linewidth}}
    \textbf{Paper} & \textbf{Dataset} & \textbf{Application} \\
    \hline
     Dynamics of Dengue with human and vector... & Mosquito density data & PDE reaction-diffusion \\
     \hline
    Detection of correlation between genotypes and env... & genome-wide datasets & INLA-SPDE \\
    \end{tabular}
\end{mdframed}

Showing 2 of 17 results for datasets in which PDEs were used, thereby finding potential for converging modeling paradigms between fields of study (e.g. dynamical models used in different fields).

\vspace{4mm}
\pagebreak

\textbf{Modality Distribution Analysis}
\begin{mdframed}[
   linewidth=2pt,
   linecolor=gray!40,
   topline=false,
   rightline=false,
   bottomline=false,
   backgroundcolor=gray!5,
   innerleftmargin=10pt,
   innerrightmargin=10pt
]
   \textit{Example: What \% of papers in astrophysics use only galaxy images, only spectra, both of them, or neither?}
   \vspace{2mm}

   \begin{tabular}{p{0.3\linewidth}|p{0.25\linewidth}|p{0.2\linewidth}}
   \textbf{Category} & \textbf{Paper Count} & \textbf{Percentage (\%)}\\
   \hline
   Neither & 901 & 57.1 \\
   Only Spectra & 558 & 35.36 \\
   Only Images & 88 & 5.58 \\
   \end{tabular}
\end{mdframed}

Analysis revealed distinct patterns in modality usage: 5.58\% of papers used only image data, 35.36\% used spectra, and 1.96\% used both, while 57.1\% used neither, potentially looking at other modalities instead, such as time-series data, which appeared in 1.22\% of the papers.

\section{Related Work}
Large language models (LLMs) \citep{brown2020languagemodelsfewshotlearners, kojima2023largelanguagemodelszeroshot, openai2024gpt4technicalreport, touvron2023llama2openfoundation, grattafiori2024llama3herdmodels} have recently demonstrated remarkable capabilities in language tasks, particularly through advances in prompting strategies \citep{brown2020languagemodelsfewshotlearners, wei2023chainofthoughtpromptingelicitsreasoning}. These advances have inspired building LLM-based pipelines to engage with complex scientific discovery tasks \citep{doi:10.1073/pnas.1914370116, ai4science2023impactlargelanguagemodels, lu2024aiscientistfullyautomated}. Despite these advancements, the problem of hallucinations in LLMs persists \citep{Huang_2024}.

To overcome this, a popular approach is to augment LLMs with unstructured, external knowledge \citep{lála2023paperqaretrievalaugmentedgenerativeagent, lewis2021retrievalaugmentedgenerationknowledgeintensivenlp}, but while RAG excels in retrieving broad context information at scale, it is limited in providing precise information and avenues for systematic analysis, which can be efficiently realized through structured knowledge representations.

Some prior work applies semistructured knowledge representations to the scientific literature, such as \citet{gu2024interestingscientificideageneration}'s SciMuse system which combines LLMs with the RAKE algorithm to extract concepts as keyword phrases. \citep{sun2024knowledgegraphastronomicalresearch} extended this to astronomy, using LLMs to extract concepts from scientific texts and construct knowledge graphs, grouping the concepts with a vector-based semantic similarity. In biomedicine, \citep{nadkarni2021scientificlanguagemodelsbiomedical} integrated domain-specific language models with knowledge graph embeddings, showing improved performance but requiring field-specific finetuning.

Our work differs from these approaches by providing a domain-agnostic framework that combines LLM-powered semantic understanding with queryable structured knowledge curation, and enhanced by graph visualizations. Unlike previous methods, our approach introduces generalizable categorization schemes that enable cross-domain concept extraction, relationship mapping, and question answering.

\section{Discussion}
Our tool enables quantitative analysis of research methodologies across scientific domains by organizing concepts into distinct categories like methods, instruments, and data modalities, allowing researchers to systematically investigate patterns that would be difficult to discover through traditional literature review or citation analysis. The combination of SQL queries and graph visualization proves especially valuable for exploring methodological connections. For instance, answering how similar mathematical models get applied across different fields is hard with existing approaches but readily solvable by our system.

While the system shows promise, its current extraction precision leaves room for improvement. At times, the LLM can struggle to distinguish specific named entities (like "Melbourne wind tunnel") from generic concepts (like "wind-tunnel data"), introducing noise into the extracted relationships. Future work could address these limitations through improved prompting strategies, post-training of models by gathering insights from domain experts, and extracting more sophisticated relationships between concepts beyond co-occurrence.

\section{Conclusion}
This work demonstrates how combining LLMs with structured knowledge representation can enable systematic analysis of scientific literature. Our four key contributions are: a domain-agnostic schema for categorizing scientific concepts, a scalable LLM-based extraction pipeline, a queryable interactive system, and informative knowledge graphs built from the extracted concepts. The results show that even with modest extraction accuracy, our approach can reveal valuable insights about cross-disciplinary connections and research evolution that would be difficult to discover through traditional literature review, opening up new possibilities for navigating scientific research.

\section*{Acknowledgments}
The computations in this work were, in part, run at facilities supported by the Scientific Computing Core at the Flatiron Institute, a division of the Simons Foundation. Polymathic AI acknowledges support provided by the Simons Foundation and Schmidt Sciences, LLC.

\bibliography{custom}

\newpage
\appendix

\section{Comparison of Keyword Extraction Methods}
\subsection{Methodology}
In this appendix we compare two approaches to scientific concept extraction: Rapid Automatic Keyword Extraction (RAKE) \citep{Rose2010AutomaticKE} and our method. RAKE is an unsupervised statistical method based on word frequency and co-occurrence, while our schema-based approach leverages Llama-3 70B for semantic understanding.
We randomly sampled 300 scientific papers (100 each from astrophysics, fluid dynamics, and evolutionary biology) from our dataset and applied RAKE extractions on the title and abstract text, performed using the \texttt{rake-nltk} library \citep{vishwas2017rakenltkpython}. The extracted keywords were then compared to those generated by our method.

\subsection{Quantitative Comparison}
Table \ref{tab:concept_comparison} reveals significant differences in the average count of extracted concepts by each method across domains for the subset of 300 papers.

\begin{table}[h]
\centering
\small
\begin{tabular}{@{}lccc@{}}
\toprule
\textbf{Domain} & \textbf{RAKE} & \textbf{Ours} & \textbf{Ratio} \\
 & \textbf{(avg)} & \textbf{(avg)} & \textbf{(avg)} \\
\midrule
Astrophysics & 69.5 & 33.8 & 2.05 \\
Fluid Dynamics & 70.3 & 32.6 & 2.15 \\
Evolutionary Biology & 67.7 & {29.7} & 2.28 \\
\hline
\textbf{Overall} & \textbf{69.2} & \textbf{32.1} & \textbf{2.16} \\
\bottomrule
\end{tabular}
\caption{Average number of extracted concepts by domain}
\label{tab:concept_comparison}
\end{table}

\begin{table}[h]
\centering
\small
\begin{tabular}{@{}lp{0.64\columnwidth}@{}}
\toprule
\textbf{Domain} & \textbf{Our method: concept types} \\
\midrule
Astrophysics & object (56.1\%), property (24.7\%), instrument (5.4\%), method (3.6\%), modality (3.5\%), model (3.1\%), task (1.5\%), field (1.0\%), dataset (0.6\%) \\
\midrule
Fluid Dynamics & object (44.7\%), property (31.0\%), method (10.0\%), model (5.3\%), modality (3.3\%), field (1.9\%), task (1.9\%), instrument (1.2\%), dataset (0.5\%) \\
\midrule
Evolutionary Biology & object (51.2\%), property (25.7\%), model (9.2\%), method (5.4\%), task (3.3\%), field (2.5\%), modality (1.7\%), dataset (0.4\%), instrument (0.2\%) \\
\bottomrule
\end{tabular}
\caption{Distribution of concept types by domain}
\label{tab:concept_types}
\end{table}

Our approach consistently extracted fewer concepts than RAKE but organized them into semantic categories that reveal their functional roles within the scientific discourse. In table \ref{tab:concept_types}, we see that across all domains, \texttt{object} was the predominant concept type, followed by \texttt{property}.

\subsection{Qualitative Analysis}
\subsubsection{Astrophysics}
\begin{tcolorbox}[
title=\textbf{A Detailed Analysis of a Magnetic Island Observed by WISPR on Parker Solar Probe},
colback=Pink,
arc=3mm,
boxrule=0.8pt,
left=2mm,
right=2mm,
top=2mm,
bottom=2mm,
enhanced,
]
\noindent \small We present the identification and physical analysis of a possible magnetic
island feature seen in white-light images observed by the Wide-field Imager for
Solar Probe (WISPR) on board the Parker Solar Probe (Parker). The island is
imaged by WISPR during Parker's second solar encounter on 2019 April 06, when
Parker was ~38 solar radii from the Sun center. We report that the average
velocity and acceleration of the feature are approximately 334 km s and -0.64 m
s-2. The kinematics of the island feature, coupled with its direction of
propagation, indicate that the island is likely entrained in the slow solar
wind. The island is elliptical in shape with a density deficit in its center,
suggesting the presence of a magnetic guide field. We argue that this feature
is consistent with the formation of this island via reconnection in the current
sheet of the streamer. The feature's aspect ratio (calculated as the ratio of
its minor to major axis) evolves from an elliptical to a more circular shape
that approximately doubles during its propagation through WISPR's field of
view. The island is not distinct in other white-light observations from the
Solar and Heliospheric Observatory (SOHO) and the Solar Terrestrial Relations
Observatory (STEREO) coronagraphs, suggesting that this is a comparatively
faint heliospheric feature and that viewing perspective and WISPR's enhanced
sensitivity are key to observing the magnetic island.
\end{tcolorbox}

\begin{tcolorbox}
[
title=\textbf{RAKE Keywords\\\footnotesize[(score, "keyword") * count]},
arc=3mm,
boxrule=0.8pt,
left=2mm,
right=2mm,
top=2mm,
bottom=2mm,
]
\noindent \small
    (17.67, "possible magnetic island feature seen"), (13.71, "solar terrestrial relations observatory"), (13.33, "comparatively faint heliospheric feature"), (9.00, "2019 april 06"), (8.71, "slow solar wind"), (8.71, "second solar encounter"), (8.71, "38 solar radii"), (8.50, "light images observed"), (8.50, "approximately 334 km"), (8.33, "magnetic guide field"), (8.00, "island via reconnection"), (6.96, "parker solar probe"), (6.00, "heliospheric observatory"), (5.33, "magnetic island"), (5.21, "solar probe"), (4.50, "light observations"), (4.50, "approximately doubles"), (4.33, "island feature"), (4.00, "viewing perspective"), (4.00, "physical analysis"), (4.00, "major axis"), (4.00, "likely entrained"), (4.00, "field imager"), (4.00, "enhanced sensitivity"), (4.00, "density deficit"), (4.00, "current sheet"), (4.00, "average velocity"), (3.75, "parker ()"), (3.50, "sun center"), (3.50, "circular shape"), (3.50, "aspect ratio"), (2.71, "solar"), (2.33, "feature") * 3, (2.00, "island") * 4, (2.00, "field"), (1.75, "parker") * 2, (1.50, "shape"), (1.50, "ratio"), (1.50, "center"), (1.00, "wispr") * 4, (1.00, "wide"), (1.00, "white") * 2, (1.00, "view"), (1.00, "suggesting") * 2, (1.00, "streamer"), (1.00, "stereo"), (1.00, "soho"), (1.00, "report"), (1.00, "propagation") * 2, (1.00, "present"), (1.00, "presence"), (1.00, "observing"), (1.00, "minor"), (1.00, "kinematics"), (1.00, "key"), (1.00, "indicate"), (1.00, "imaged"), (1.00, "identification"), (1.00, "formation"), (1.00, "evolves"), (1.00, "elliptical") * 2, (1.00, "distinct"), (1.00, "direction"), (1.00, "coupled"), (1.00, "coronagraphs"), (1.00, "consistent"), (1.00, "calculated"), (1.00, "board"), (1.00, "argue"), (1.00, "acceleration"), (1.00, "64"), (1.00, "2"), (1.00, "0")
\end{tcolorbox}

\begin{tcolorbox}
[
title=\textbf{Concepts (by type):}
]
\noindent \small
\begin{itemize}
\item \textbf{object}: Parker Solar Probe, magnetic island feature, white-light images, island, Parker, Sun, feature, island feature, slow solar wind, streamer, WISPR's field of view, magnetic island, heliospheric feature
\item \textbf{instrument}: WISPR, Wide-field Imager for Solar Probe (WISPR), Parker Solar Probe (Parker), Solar and Heliospheric Observatory (SOHO), Solar Terrestrial Relations Observatory (STEREO)
\item \textbf{property}: solar radii, velocity, acceleration, density deficit, shape, aspect ratio, faint
\end{itemize}
\end{tcolorbox}

The RAKE extraction produces a flat list of keywords with associated scores. In contrast, our approach organizes concepts by semantic role, such as instruments (WISPR, SOHO), objects (Sun, slow solar wind), and their properties (solar radii, acceleration).

\subsubsection{Fluid Dynamics}
\begin{tcolorbox}[
title=\textbf{Approximation of sea surface velocity field by fitting surrogate two-dimensional flow to scattered measurements},
colback=Pink,
arc=3mm,
boxrule=0.8pt,
left=2mm,
right=2mm,
top=2mm,
bottom=2mm,
enhanced,
]
\noindent
\small In this paper, a rapid approximation method is introduced to estimate the sea
surface velocity field based on scattered measurements. The method uses a
simplified two-dimensional flow model as a surrogate model, which mimics the
real submesoscale flow. The proposed approach treats the interpolation of the
flow velocities as an optimization problem, aiming to fit the flow model to the
scattered measurements. To ensure consistency between the simulated velocity
field and the measured values, the boundary conditions in the numerical
simulations are adjusted during the optimization process. Additionally, the
relevance of quantity and quality of the scattered measurements is assessed,
emphasizing the importance of the measurement locations within the domain as
well as explaining how these measurements contribute to the accuracy and
reliability of the sea surface velocity field approximation. The proposed
methodology has been successfully tested in both synthetic and real-world
scenarios, leveraging measurements obtained from Global Positioning System
(GPS) drifters and high-frequency (HF) radar systems. The adaptability of this
approach for different domains, measurement types, and conditions implies that
it is suitable for real-world submesoscale scenarios where only an
approximation of the sea surface velocity field is sufficient. 
\end{tcolorbox}

\begin{tcolorbox}
[
title=\textbf{RAKE Keywords\\\footnotesize[(score, "keyword") * count]},
arc=3mm,
boxrule=0.8pt,
left=2mm,
right=2mm,
top=2mm,
bottom=2mm,
]
\noindent \small (22.83, "sea surface velocity field based"), (20.83, "sea surface velocity field approximation"), (17.83, "sea surface velocity field"), (11.50, "simulated velocity field"), (9.00, "global positioning system"), (8.50, "rapid approximation method"), (8.50, "measurement locations within"), (8.20, "leveraging measurements obtained"), (8.00, "world submesoscale scenarios"), (7.83, "dimensional flow model"), (7.50, "proposed approach treats"), (7.17, "real submesoscale flow"), (5.00, "world scenarios"), (4.83, "flow model"), (4.50, "proposed methodology"), (4.50, "method uses"), (4.50, "measurement types"), (4.50, "flow velocities"), (4.33, "surrogate model"), (4.20, "scattered measurements") * 3, (4.20, "measurements contribute"), (4.00, "successfully tested"), (4.00, "simplified two"), (4.00, "radar systems"), (4.00, "optimization process"), (4.00, "optimization problem"), (4.00, "numerical simulations"), (4.00, "measured values"), (4.00, "ensure consistency"), (4.00, "different domains"), (4.00, "conditions implies"), (4.00, "boundary conditions"), (3.00, "approximation"), (2.00, "approach"), (1.67, "real") * 2, (1.00, "well"), (1.00, "synthetic"), (1.00, "suitable"), (1.00, "sufficient"), (1.00, "reliability"), (1.00, "relevance"), (1.00, "quantity"), (1.00, "quality"), (1.00, "paper"), (1.00, "mimics"), (1.00, "introduced"), (1.00, "interpolation"), (1.00, "importance"), (1.00, "high"), (1.00, "hf"), (1.00, "gps"), (1.00, "frequency"), (1.00, "fit"), (1.00, "explaining"), (1.00, "estimate"), (1.00, "emphasizing"), (1.00, "drifters"), (1.00, "domain"), (1.00, "assessed"), (1.00, "aiming"), (1.00, "adjusted"), (1.00, "additionally"), (1.00, "adaptability"), (1.00, "accuracy")
\end{tcolorbox}

\begin{tcolorbox}
[
title=\textbf{Concepts (by type):}
]
\noindent \small
\begin{itemize}
\item \textbf{object}: sea surface velocity field, surrogate two-dimensional flow, measurements, submesoscale flow, flow velocities, velocity field, measurement locations, domain, domains
\item \textbf{method}: fitting, rapid approximation method, simplified two-dimensional flow model, optimization, optimization process, methodology
\item \textbf{property}: velocity, measured values, quantity, quality, accuracy, reliability, adaptability
\item \textbf{model}: two-dimensional flow model, flow model
\item \textbf{task}: interpolation
\item \textbf{instrument}: Global Positioning System (GPS) drifters, high-frequency (HF) radar systems
\item \textbf{modality}: high-frequency, measurement types
\end{itemize}
\end{tcolorbox}

Our method's extraction effectively distinguishes between physical objects of study (porous media), models (two-dimensional flow model), methodological approaches (rapid approximation method), and instruments (Global Positioning System (GPS) drifters).

\subsubsection{Evolutionary Biology}
\begin{tcolorbox}[
title=\textbf{Complexity-stability relationships in disordered dynamical systems},
colback=Pink,
arc=3mm,
boxrule=0.8pt,
left=2mm,
right=2mm,
top=2mm,
bottom=2mm,
enhanced,
]
\noindent
\small Robert May famously used random matrix theory to predict that large, complex
systems cannot admit stable fixed points. However, this general conclusion is
not always supported by empirical observation: from cells to biomes, biological
systems are large, complex and, often, stable. In this paper, we revisit May's
argument in light of recent developments in both ecology and random matrix
theory. Using a non-linear generalization of the competitive Lotka-Volterra
model, we show that there are, in fact, two kinds of complexity-stability
relationships in disordered dynamical systems: if self-interactions grow faster
with density than cross-interactions, complexity is destabilizing; but if
cross-interactions grow faster than self-interactions, complexity is
stabilizing.
\end{tcolorbox}

\begin{tcolorbox}
[
title=\textbf{RAKE Keywords\\\footnotesize[(score, "keyword") * count]},
arc=3mm,
boxrule=0.8pt,
left=2mm,
right=2mm,
top=2mm,
bottom=2mm,
] 
\noindent \small
(40.50, "robert may famously used random matrix theory"), (40.00, "complex systems cannot admit stable fixed points"), (15.00, "random matrix theory"), (10.00, "disordered dynamical systems"), (8.00, "interactions grow faster") * 2, (6.50, "revisit may"), (6.00, "biological systems"), (4.00, "volterra model"), (4.00, "two kinds"), (4.00, "stable"), (4.00, "stability relationships"), (4.00, "recent developments"), (4.00, "linear generalization"), (4.00, "general conclusion"), (4.00, "empirical observation"), (4.00, "complex"), (4.00, "competitive lotka"), (4.00, "always supported"), (2.00, "interactions") * 2, (1.00, "using"), (1.00, "stabilizing"), (1.00, "show"), (1.00, "self") * 2, (1.00, "predict"), (1.00, "paper"), (1.00, "often"), (1.00, "non"), (1.00, "light"), (1.00, "large") * 2, (1.00, "however"), (1.00, "fact"), (1.00, "ecology"), (1.00, "destabilizing"), (1.00, "density"), (1.00, "cross") * 2, (1.00, "complexity") * 3, (1.00, "cells"), (1.00, "biomes"), (1.00, "argument")
\end{tcolorbox}

\begin{tcolorbox}
[
title=\textbf{Concepts (by type):}
]
\noindent \small
\begin{itemize}
\item \textbf{object}: disordered dynamical systems, systems, cells, biomes, biological systems
\item \textbf{property}: complexity-stability relationships, large, complex, stable, complexity-stability, density
\item \textbf{method}: random matrix theory
\item \textbf{field}: ecology
\item \textbf{model}: non-linear generalization of the competitive Lotka-Volterra model
\end{itemize}
\end{tcolorbox}
Our method's extraction captures biological entities (cells, biomes, systems) and their properties, while also identifying the specific models and methods used. 

\subsection{Insights From The Comparison}


RAKE is efficient, language-independent, and quantitatively ranks keyword importance, but it lacks semantic depth, extracts fragmented phrases, and does not distinguish between concept types. Our method, while computationally more demanding and occasionally prone to concept hallucination, provides structured semantic categorization, generates coherent concepts, and captures domain-specific nuances more effectively.

Both methods introduce some noise, though RAKE produces significantly more. The structured semantic representation in our approach offers a more meaningful and organized summary compared to RAKE’s flat keyword list, making it more useful for domain experts.

\end{document}